\documentclass[10pt,twocolumn,letterpaper]{article}

\usepackage[pagenumbers]{wacv} 

\usepackage{graphicx}
\usepackage{amsmath}
\usepackage{amssymb}
\usepackage{booktabs}
\usepackage[dvipsnames]{xcolor}
\usepackage{enumitem}

\newcommand{\supplementary}{\hyperref[sec:appendix]{Appendix}}
\DeclareRobustCommand*{\RaiseBoxByDepth}{%
    \raisebox{\depth}%
}

\usepackage[pagebackref,breaklinks,colorlinks]{hyperref}

\usepackage[capitalize]{cleveref}
\crefname{section}{Sec.}{Secs.}
\Crefname{section}{Section}{Sections}
\Crefname{table}{Table}{Tables}
\crefname{table}{Tab.}{Tabs.}

\def\wacvPaperID{*****} 
\def\confName{*****}
\def\confYear{*****}

\begin{document}

\title{Learning to Visually Connect Actions and their Effects}

\author{Paritosh Parmar \hspace{1cm} Eric Peh \hspace{1cm} Basura Fernando\\
\small{Institute of High-Performance Computing, Agency for Science, Technology and Research, Singapore}\\
}

\maketitle

\begin{abstract}
We introduce the novel concept of visually Connecting Actions and Their Effects (CATE) in video understanding. CATE can have applications in areas like task planning and learning from demonstration. We identify and explore two different aspects of the concept of CATE: Action Selection (AS) and Effect-Affinity Assessment (EAA), where video understanding models connect actions and effects at semantic and fine-grained levels, respectively. We design various baseline models for AS and EAA. Despite the intuitive nature of the task, we observe that models struggle, and humans outperform them by a large margin. Our experiments show that in solving AS \& EAA, models learn intuitive properties like object tracking and pose encoding without explicit supervision. We demonstrate that CATE can be an effective self-supervised task for learning video representations from unlabeled videos. The study aims to showcase the fundamental nature and versatility of CATE, with the hope of inspiring advanced formulations and models. \textcolor{RubineRed}{Dataset and Code will be publicly released.}
\end{abstract}
\section{Introduction}
\label{sec:introduction}

\begin{figure}
    \centering
    \includegraphics[width=0.9\columnwidth]{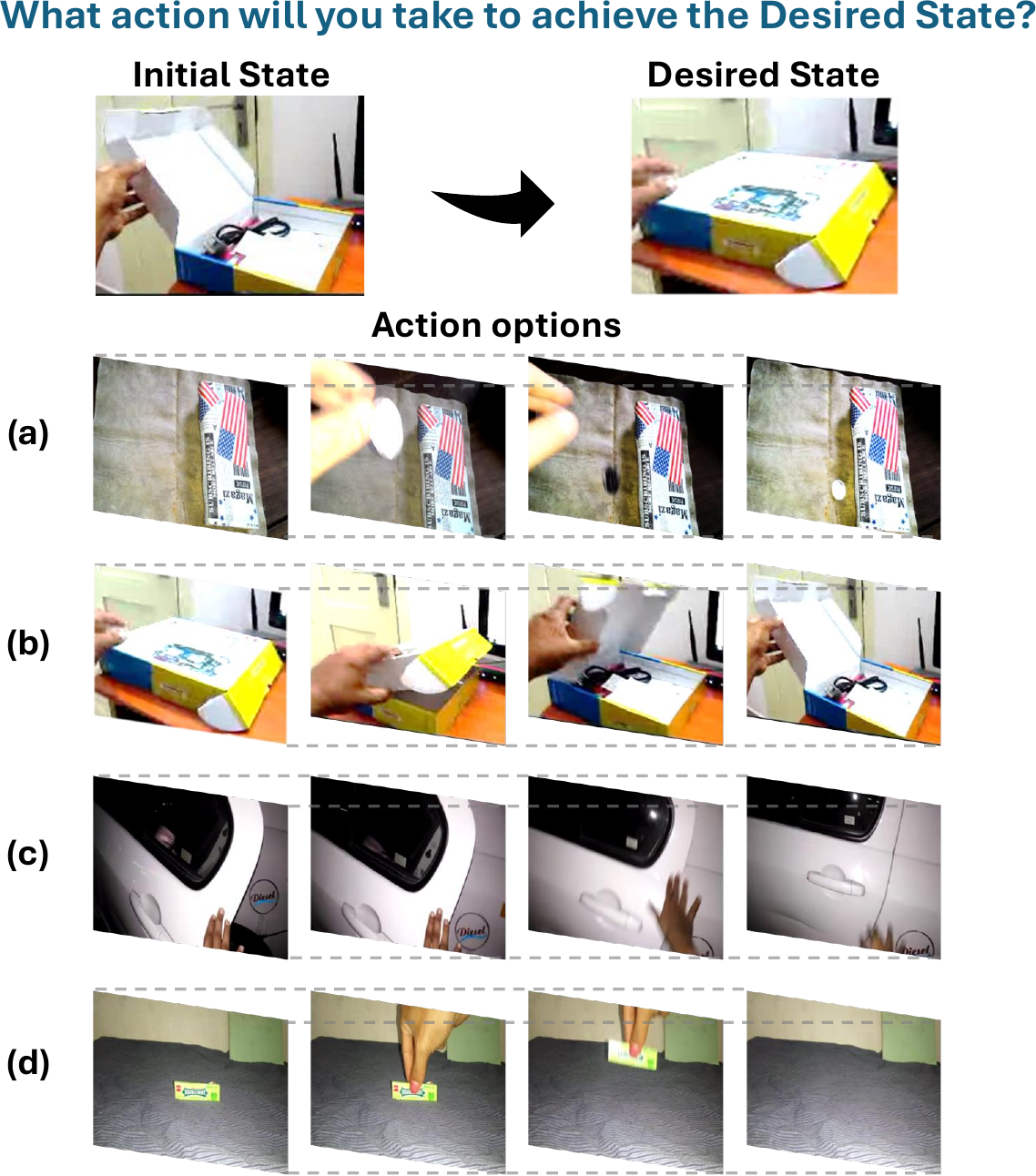}
    \caption{Ability to select the action to carry out to achieve the desired results is so effectively and widely used by humans, that it has become second nature to us---we use it without even realizing we are using. But can the video understanding models do the same thing? \RaiseBoxByDepth{\rotatebox{180}{Correct answer is (c), not (b).}}}
    \label{fig:concept}
\end{figure}

Actions can transform systems, scenes, and environments. Humans possess the remarkable ability to \underline{c}onnect \underline{a}ctions with \underline{t}heir corresponding \underline{e}ffects (CATE) and vice versa based on visual information. Given an initial visual state\footnote{In general, the state of the system before the action is applied is termed as initial state, and the state of the system after the action is applied is termed the final or the desired state.}, we can select an appropriate action and execute it to achieve the desired state/outcome, \eg, as in \autoref{fig:concept}. Conversely, given an initial state, when an action is applied, we can imagine the outcome or the resultant state of the system---for instance, consider a scenario where a person is holding an object; if they release their grip, we intuitively know that the object will inevitably fall to the ground. 

This bidirectional understanding of causal mechanisms is crucial as it allows humans to plan and execute goal-directed actions, anticipate the consequences, make informed decisions, and adjust their behavior. Integrating visual information with motor actions helps individuals build a coherent and adaptive representation of the world, enabling effective navigation of their environment. Incorporating such capabilities into video understanding systems and autonomous robots is essential, with applications in task planning and learning from demonstration. CATE can enable practical applications like: \textbf{1)} \textit{Robots} can learn to perform complex tasks by understanding the relationships between their actions \& the resulting effects in the environment; \textbf{2)} \textit{Sports AI coach \& rehab} by analyzing athletes' current movements \& ideal movements, AI system can suggest personalized training plans; \textbf{3)} \textit{Education}: Students can interact with CATE-powered simulations to understand how different actions lead to specific outcomes in physics, biomechanics, engineering. 

Despite its fundamental role in human cognition, linking actions to their physical effects is largely unexplored in video understanding. To address this, we introduce the concept of visually connecting actions and effects. Our work contributes to this understanding by investigating two key aspects of \textbf{CATE}: 1) \textbf{Action Selection} and 2) \textbf{Effect-Affinity Assessment}, offering distinct perspectives on connecting actions to effects in visual scenes.

Connecting actions and their effects can be approached from various perspectives, leading to different task formulations. For instance, actions and effects could be linked at a \textit{semantic} level. Following this perspective, we introduce the task of \textbf{Action Selection}, where video understanding models are presented with a pair of initial and final visual states of a scene, and they must select the correct action from a set of options that transitions the system from the initial state to the final or desired state (ref. \autoref{fig:concept}). While Action Selection may seem an easy and intuitive task for humans, we found it to be quite challenging for various state-of-the-art video understanding models. To address this, we propose several baseline frameworks that improve upon the current state-of-the-art models, but there remains a significant performance gap between humans and machines in this task.

Now, let us take a look at connecting actions and effects at a \textit{finer} granularity. To achieve this, we introduce the task of \textbf{Effect-Affinity Assessment}, where a video understanding model is tasked with inferring how closely or directly related is an effect to an action (\autoref{fig:concept}, \autoref{fig:ssl_action_spec}). So, while Action Selection requires connecting actions and effects at a \textit{semantic} level with \textit{coarser} granularity, Effect-Affinity Assessment involves discerning at a \textit{finer} resolution.

Interestingly, we observe that different formulations of CATE learn representations that capture \textbf{intuitive properties} of actions. For example, the \textit{Action Selection} model learns to \textit{track objects} without any explicit supervision (see \autoref{fig:attn_viz}); while \textit{Effect-Affinity Assessment} representations learn to encode fine-grained details like the \textit{pose of humans} without explicit supervision (see \autoref{fig:res_pose_ret_explained}). We also observe that connecting actions and effects can serve as an effective \textbf{self-supervised pretext task} to learn generic video representations from unlabeled videos. 

We aim to lay a foundation for future work by exploring various aspects of CATE, devising task formulations, \& developing models. Our study underscores CATE's essential role in \textit{learning}, highlighting its \textit{flexibility} across contexts \& \textit{versatility} across domains. This paves the way for advanced \textit{models} with broader \textit{applications} \& improved \textit{performance}.

The paper is organized as follows: \S\ref{sec:related} reviews \textit{related work}. \S\ref{sec:task_formulation_1} discusses \textit{Action Selection} task, dataset \& baseline model. \S\ref{sec:task_formulation_2} shows that \textit{useful representations emerge} from Action Selection. \S\ref{sec:task_formulation_3} covers \textit{Effect-Affinity Assessment} \& its model. \S\ref{sec:experiments} \textit{benchmarks} baselines on CATE tasks \& \textit{examines} learnt representations. \S\ref{sec:conclusion} \textit{concludes} the paper.
\section{Related Work}
\label{sec:related}
\noindent\textbf{Video understanding tasks.} Video understanding is a mature branch of computer vision addressing action analysis problems such action recognition \cite{ucf101, carreira2017quo, ssv2, c3d, videoswin}, action segmentation/localization \cite{kalogeiton2017action, ding2023temporal, ghoddoosian2022hierarchical, rahaman2022generalized}, and action quality assessment \cite{aqa7, fitnessaqa, yao2020video, bai2022action}. Tremendous progress has been made in these directions, but connecting actions \& their effects largely remains to be addressed. Towards that end, in this work, we study linking actions \& their effects and study consequences of the same. We make use of the techniques \& insights gleaned from existing video understanding tasks towards the problem of connecting actions \& effects.

\noindent\textbf{Viewing actions as transformations.} Actions can be viewed as transformations or agents of transformations occurring in the scene or an environment \cite{wang2016actions, look4change, nagarajan2018attributes}. Based on this, Wang \etal \cite{wang2016actions}, proposed to model actions as fixed class-specific transformation matrices, which when applied to a precondition would produce the effect. Soucek \etal \cite{look4change} proposed to learn action-specific models to localize states and actions. Similar to actions, Nagarajan and Grauman \cite{nagarajan2018attributes} view object properties as operators on objects. However, these works either focus on learning a rigid set of class-specific transformation matrices \cite{wang2016actions}, or learning only action-specific models \cite{look4change} or limited to static images \cite{nagarajan2018attributes, look4change} and do not focus on studying action dynamics and their effects. Our work studies linking action dynamics and effects in much more depth and breadth. Our task also demonstrates wider applicability. We further take inspiration from prior work \& develop their dynamic counterparts which can generate transformations on the fly from actions \& that way help in connecting actions \& their effects. 

\noindent\textbf{Synthetic datasets.} 
Synthetic datasets simplify obtaining ground-truth for object properties and positions \cite{johnson2017clevr, girdhar2019cater} and are useful for studying state changes \cite{trancenet}. However, these methods are less applicable to real-world scenarios, which lack control over state and action attributes. Our work uses real-world datasets and introduces a cross-sample setting, akin to View transformation \cite{trancenet} but more challenging due to varied factors. Unlike synthetic datasets, our experiments involve human actions in real-world contexts.

\noindent\textbf{Modeling state-changes.} 
Computing state changes from initial \& final states is addressed in works like \cite{park2019robust, jhamtani2018learning, kim2021agnostic}. While this is crucial to our problem, these works do not link state changes to the actions causing them, unlike our work.

\noindent\textbf{Assorted work on visual reasoning.} 
Liang \etal \cite{Liang_2022_CVPR} address generating likely explanations for partial observations, which is related to our work on linking initial and final states. However, they focus on higher-level explanations, while we focus on foundational connections between action mechanics and effects. Other reasoning problems introduced recently \cite{wu2021star, hessel2022abduction} are orthogonal to ours.
\section{Problem of Action Selection}
\label{sec:task_formulation_1}
Given the initial \& desired final states, the problem of Action Selection involves video understanding models to \textit{select} the \textit{correct} action instance that can \textit{achieve} the \textit{desired} state. In practical scenarios, we have initial state, but the final desired state may not be directly observable but exists as a mental representation (or other forms such as a sketch). However, to create a simplified/proof-of-concept version of the task, we assume access to the initial \& final states in the form of images. This approach helps \textit{isolate} \& \textit{focus} on the core challenge of understanding \textit{causal relationships} between actions \& outcomes \textit{without} the additional complexity of environmental modeling.

\noindent\textbf{Task Design.} We cast the problem of Action Selection as a \textit{multiple-choice question answering}; and provide four action-video choices for answers (similar to ~\autoref{fig:concept}---we have shown only three options there for simplicity). From a total of four choices/options, only one is correct. The task is to select the correct action-video. The multiple-choice format for the Action Selection task is not only \textit{practical} and reflective of both \textit{human cognitive processes} and real-world \textit{autonomous agent operational scenarios}, but it also aids in simplifying and \textit{standardizing} the learning and evaluation processes. In practical scenarios \textit{more than four} action choices/options are expected. However, as we will see in experiments, the performance of models was already quite low even with just 4 options; with more number of options the performance is only expected to go down from there. Thus, we did not consider more options in this pilot study.

Looking ahead, the Action Selection task has the potential to enable autonomous agents and robots to learn action selection from \textit{human demonstrations} found in the existing \textit{large-scale video datasets}, as discussed in \autoref{sec:CATE_dataset}.

\noindent\textbf{Constructing \textit{Cross-Sample} question-answer sets.} To form a question, we divide a video instance of a human performing an action into the \textit{initial state} of the scene (\textit{starting} few frames), \& the resultant \textit{final state} (\textit{last} few frames). The remaining \textit{middle} frames after keeping a certain margin from the initial \& final state frames become the correct \textit{action} option. However, the correct answer option can be drawn in two ways: \textbf{1)} \textit{same-sample setting}: from the same video clip as the states as explained earlier; or \textbf{2)} \textit{cross-sample setting}: from an altogether different instance, but the action category remains the same as the video from which states are drawn. \textit{Cross-sample} setting can be viewed as a very \textit{strong data augmentation} because essentially all factors such as \textit{background}, \textit{colors}, \textit{viewing angle}, \textit{actors}, \& \textit{scene setup} may be \textit{different} from state-video, while \textit{semantic action class} remains the \textit{same}. This setup enables the \textit{isolation} of \textit{action semantics} from \textit{irrelevant} features of the scene \& demands to \textit{focus} on the \textit{relevant} parts of the video, thereby helping connect actions \& their effects effectively.

\subsection{Cross-Sample Action Selection Dataset}
\label{sec:CATE_dataset}
We introduce a novel \textbf{dataset} to support our proposed \textit{Action Selection} task. Each sample in our dataset consists of the ``\textit{question}" (pair of \textit{initial} and \textit{desired} states), and the corresponding a \textbf{1} \textit{correct} answer and \textbf{3} \textit{incorrect} answers.

\noindent\textbf{Dataset sources.} To ensure a diverse dataset, we chose large-scale video datasets with numerous action classes involving significant state changes: 1) \textbf{SSv2} dataset \cite{ssv2} encompasses a diverse range of actions, including \textit{physical movements}, \textit{object handling}, \textit{transformations}, and \textit{interactions}. It includes actions such as \textit{lifting}, \textit{pushing}, \textit{pouring}, \textit{stacking}, and more, involving the \textit{manipulation} and \textit{alteration} of \textit{objects}. 2) \textbf{COIN} dataset \cite{coin} focuses on \textit{specific tasks} and \textit{procedures} related to \textit{assembling}, \textit{installing}, \textit{building}, or \textit{manipulating} various \textit{objects} and \textit{materials}. It includes step-by-step (we consider a \textit{step} as an \textit{action}) instructions for tasks like \textit{assembly}, \textit{installation}, \textit{cooking}, and \textit{maintenance}, emphasizing precise and ordered procedures. 

\noindent\textbf{Selecting action classes.} 
Not all action classes from SSv2 and COIN involve \textit{significant state changes}, making them \textit{less} suitable for the Action Selection task. For instance, classes like ``\textit{approaching something with camera}" were \textit{discarded}. We \textit{manually} filtered usable action classes by evaluating their \textit{names} and \textit{inspecting} randomly sampled videos to verify significant state changes. This process yielded a total of \textbf{204} \textit{usable} action classes.  \textit{No overlap} of classes was ensured \textit{manually} to avoid model confusion. Further details on dataset are provided in \supplementary.

\noindent\textbf{Temporal annotations for states and actions.} 
To \textit{capture} states and actions \textit{effectively}, we use the \textit{first} \& \textit{last} \textbf{10\%} of all video frames as the \textit{initial} and \textit{final} states, respectively, with the \textit{middle} \textbf{80\%} representing the driving \textit{action}. Using \textit{percentages} rather than \textit{fixed} frame counts, this method accommodates \textit{varying} video lengths. To address potential \textit{noise} in crowd-sourced datasets like SSv2, where clips may include \textit{irrelevant} or \textit{shaky} footage, we detect \textbf{objects} and \textbf{hands}, retaining footage only when they are \textit{detected}. This enhances dataset quality, as shown in \supplementary.

\noindent\textbf{Constructing quadruplets.} We follow the \textbf{cross-sample} setting as mentioned in \autoref{sec:task_formulation_1} to construct quadruplets of \textit{question}, \textit{correct answer}, \& \textit{incorrect answer} options. In total, we prepare \textbf{2} million question-answer sets (\textbf{1}M from each SSv2, COIN datasets). Number of Train/Val/Test questions: \textbf{1.4}M, \textbf{0.3}M, \textbf{0.3}M.

\noindent\textbf{Causal-Confusion Hard Counterfactuals.} 
Incorrect action options can be created by drawing them from different action categories. However, we introduce a \textit{harder-to-distinguish} strategy called \textit{Causal-Confusion} hard counterfactuals. This involves applying one of the following transformations to the correct action option: \textbf{1)} \textit{Temporal-reversal}: Temporally reverse the action clip (\eg, option (b) in \autoref{fig:concept}). \textbf{2)} \textit{Static}: Repeat the first action frame, meaning no real action occurs. \textbf{3)} \textit{Horizontal flipping}: Actions moving left to right will have opposite meanings. These transformations generate samples that \textit{look identical} to the correct action class but \textit{differ} in causal factors. We \textit{manually} shortlisted \textit{appropriate} augmentations for each action, ensuring that \textit{unsuitable} augmentations are \textit{not} applied to any action. The \textit{augmentation list} will be \textcolor{RubineRed}{released} along with the dataset to aid future work. This strategy encourages the video understanding model to \textit{focus} on causal factors while learning to be \textit{invariant} to irrelevant factors such as background and appearance.

\begin{figure}
\small
\centering
\includegraphics[width=0.9\columnwidth]{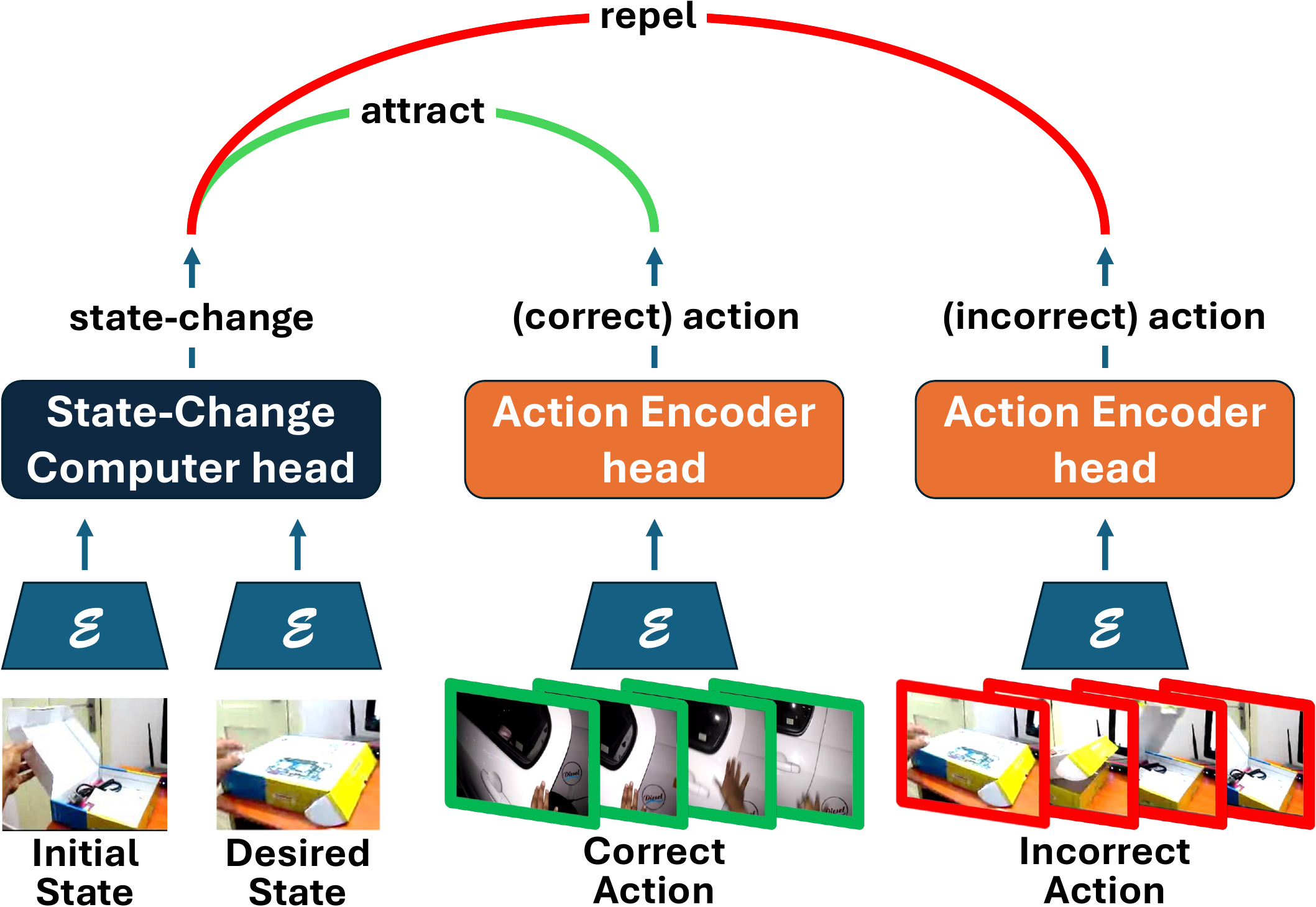}
\caption{\textbf{Cross-sample Analogical reasoning model.} Here, we have shown only one incorrect action; in practice, we use all the incorrect options for counterfactual reasoning.}
\label{fig:model_AR}
\end{figure}

\subsection{Cross-Sample Analogical Reasoning Model for Action Selection}
Inspired by \cite{nattkemper2004cognitive, herd2021neural}, we develop a human cognition-inspired baseline model for Action Selection---\textbf{Analogical Reasoning model}, detailed below. Model is shown in \autoref{fig:model_AR}.
\noindent\textbf{Model overview.} The \textit{goal} of the Analogical Reasoning model is to \textit{select} the \textit{correct} action that can help achieve the desired state starting from the initial state. During \textbf{training}, the model learns to map \textit{state changes} \& \textit{actions} into a \textbf{joint} space using \textit{contrastive} learning \cite{chen2020simple}. This involves \textit{maximizing} the (cosine-)\textit{similarity} between the representations of state changes \& \textit{correct} actions while \textit{minimizing} the similarity between the representations of state changes \& \textit{incorrect} actions. During \textbf{inference}, the alignment between state change \& action representations is measured in this joint space. The action with the \textit{best} alignment (cosine similarity) to the state change feature vector is selected as the \textit{correct} action from the available action options to achieve the desired state from the initial state.

\noindent\textbf{Framework.} Actions are represented as video clips, while states by very short clips. Unless specified otherwise, we use a backbone model ($\mathcal{E}$) pretrained on a large-scale action recognition dataset to extract features for actions and states. Subsequently, various modules mentioned in the following operate on these features. The framework consists of a \textbf{state change computer} (SCC) head, and an action encoder head. SCC head takes in the scene information abstracted by backbone from initial and final states; and \textit{analyzes} the two sets of features and computes the \textit{state-change} occurring between the two states. To do this, the SCC head \textit{concatenates} the two sets of features and processes them through a multi-layer perceptron, yielding a \textit{compact} feature vector representing the state change. Operations like \textit{addition} and \textit{subtraction} can be alternatives to our \textit{deep} SCC. However, we hypothesize that such operations are more likely to work where camera motion is \textit{absent} or minimal. In the presence of camera motion, the natural pixel-to-pixel correspondence \textit{no} longer holds, and such \textit{hard-coded} operations would \textit{underperform}, while more \textit{flexible} solutions like our \textit{deep} SCC would be better able to handle such \textit{noisy} cases. The \textbf{Action encoder} extracts information from action clip. We hypothesize that the SCC \textbf{focuses} on \textit{object placement} and \textit{pose}; while the Action Encoder focuses on \textit{motion} and \textit{tracks objects} as shown in \autoref{fig:attn_viz}.

\noindent\textbf{Training.} We use \textit{counterfactual} reasoning \cite{chen2020simple}, maximizing similarity between state changes \& correct actions while minimizing it for incorrect actions. This approach \textit{outperformed} only maximizing similarity with correct actions \textit{without} discriminating incorrect actions in our experiments.

\noindent\textbf{Optimization strategy.} We found that \textit{alternating} between optimizing state- \& action-heads improves performance.

We hypothesize that this framework involves the SCC learning to \textit{imagine} actions from states and making \textit{analogies} between \textit{imagined} and \textit{actual} actions. Thus, we term this model the \textit{Analogical Reasoning model}.
\section{Emergent Representations from using \\Action Selection as Self-Supervision}
\label{sec:task_formulation_2}

The concept of connecting actions and their effects is closely related to the idea of the \textit{sensorimotor loop} and the reciprocal relationship between \textit{vision} and \textit{movement/action} in human development \cite{gibson2014ecological, gibson2000ecological}. This reciprocal relationship involves learning via \textit{self-supervision}. Similarly, we hypothesize that useful action representations can be learnt by deep video understanding models via learning to visually \textit{connect actions} and their \textit{effects} in a \textit{self-supervised} way from \textit{unlabeled} videos. Towards that end, in this section, we show how CATE can be a useful source of \textit{intrinsic self-supervision} for visual feature learning.

\begin{figure}[!t]
\small
\centering
\includegraphics[width=0.95\columnwidth]{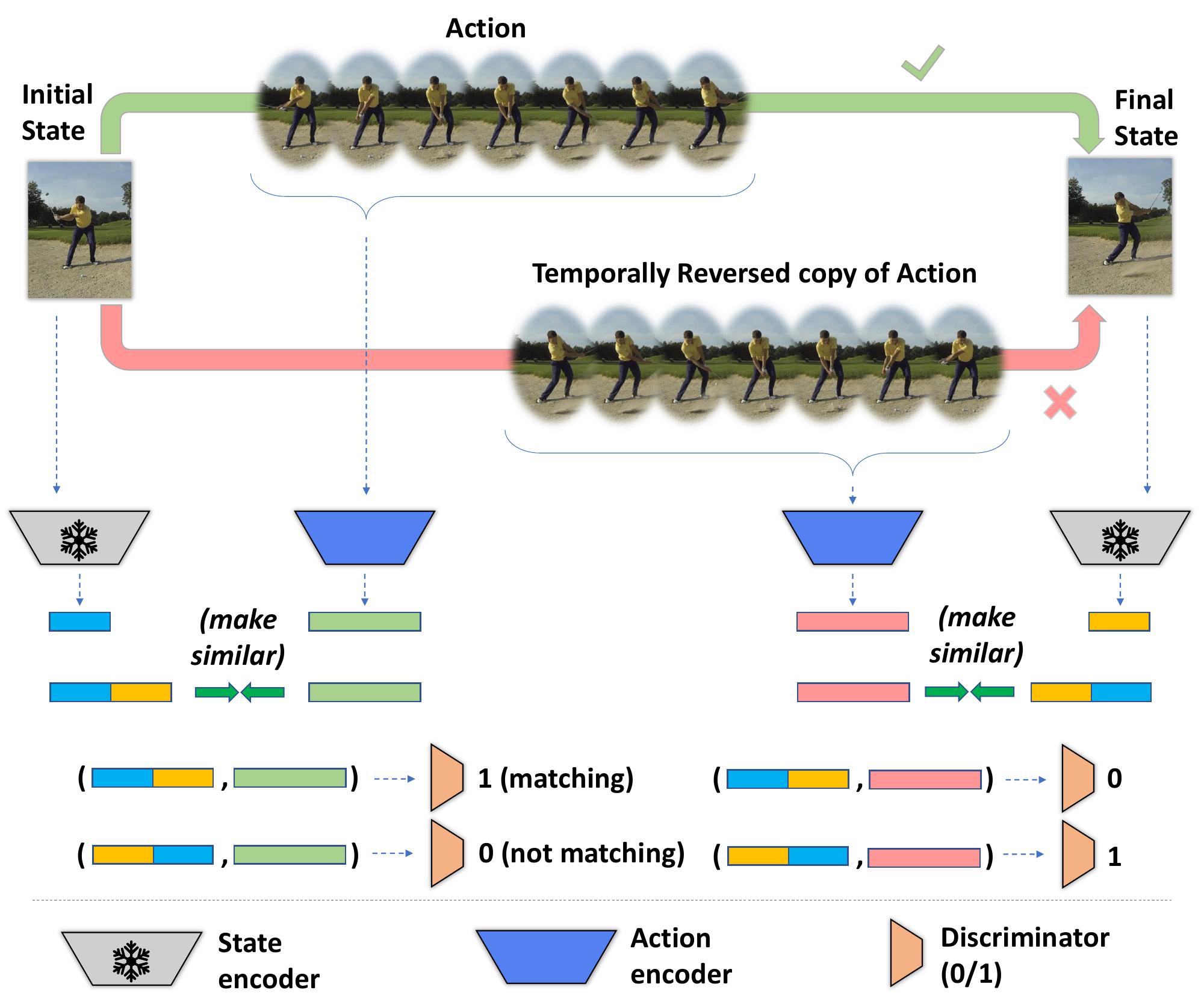}
\caption{\textbf{Self-supervised pretext task based on connecting actions and effects.} \textit{Please zoom-in to view better}. Applying action can bring the scene from its initial state to the final state, while applying a temporally reversed version of it cannot. State-encoders are frozen; state-encoders and discriminator can be discarded after training---only the action encoders are retained for further usage.}
\label{fig:ssl_action_sel}
\end{figure}

\noindent\textbf{Self-Supervised Task Design.} Given an unlabeled video of some human action, the \textit{first} and \textit{last} frames are chosen to be representatives of the \textit{initial} and the \textit{final} states of a scene. The remaining \textit{middle} portion of the video is considered as the \textit{action}(-clip). Now, we design a \textit{self-supervised task} based on the observation that applying action to the initial state would yield the final state; however, applying a \textbf{temporally-reversed} version of the same action to the initial state would \textit{not} yield the given final state. In other words, the video understanding model is tasked with selecting the action which can take the scene from its initial state to final state---\textit{forward} version of the action would, while \textit{temporally-reversed} would \textit{not}. We hypothesize in the process of solving this task, the model can learn \textit{useful video representations}. This task has some resemblance with works like \cite{misra2016shuffle, fernando2017self}, however, these works randomly shuffle frames, while our task operates specifically on the \{initial state, action, final state\} triplet. With random shuffling, potentially \textit{low-level cues} could be leveraged; while our formulation offers connecting and discerning actions and effects at a \textit{semantically} higher level. The concept along with implementation framework is shown in \autoref{fig:ssl_action_sel}. Note that the framework is a slightly modified version of Analogical Reasoning model from \S~\ref{sec:task_formulation_1} and explained further below. 

\noindent\textbf{Framework.} First, let us introduce two \textit{ordered} paired representations: \textbf{1)} \textit{forward state-change}: \{initial state representation, final state representation\}---order-preserving concatenation of the state representations; \textbf{2)} \textit{backward state-change}: \{final state representation, initial state representation\}. To avoid feature collapse, state representations are extracted using a SSL-pretrained feature extractor \cite{zbontar2021barlow}. Note that, using SSL-pretrained feature extractor for downstream SSL methods is an accepted common practice as done in, \eg, \cite{Xiao_2019_ICCV, sun2019learning, he2020momentum}. 
During \textbf{training}, the \textit{Action Encoder} learns to makes \textit{forward-action} representation \textit{similar} to \textit{forward state-change}, and \textit{backward-action} representation \textit{similar} to \textit{backward state-change}. Once trained, the \textit{Action Encoder} is used for downstream tasks like \textit{action recognition}, and the discriminator is discarded.
\section{Problem of Effect-Affinity Assessment}
\label{sec:task_formulation_3}
\begin{figure}[!t]
\small
\centering
\includegraphics[width=0.95\columnwidth]{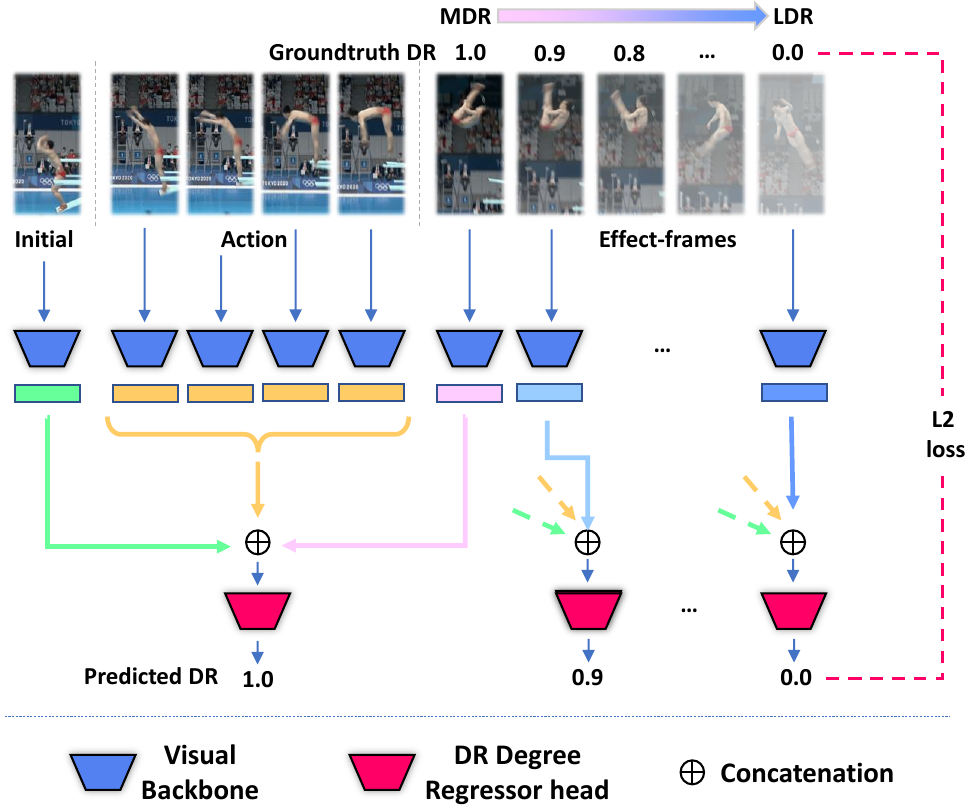}
\caption{\textbf{Self-supervised Effect-Affinity Assessment.} \textit{Zoom-in to view better.} MDR: more directly related effect; LDR: less directly related effect. Degree of how directly related an effect is given on a scale of 1.0 to 0.0 (written above Effect-frames). Farther the effect-frame from the action, lesser directly related it is to the action. Here we have shown cropped video frames to focus on the divers' pose; in practice, we use the entire frame, \& the network learns to focus on the diver through Effect-Affinity Assessment-based self-supervision.
}
\label{fig:ssl_action_spec}
\end{figure}
Effect-Affinity Assessment aims to achieve a more \textit{nuanced} understanding by focusing on discerning subtle effects in close proximity and determining the ``\textit{distance}" of an effect-frame from the action applied to an initial state. This task is formulated based on the observation that, when a short action is applied to an initial state, the frame immediately following the completion of the short action represents the most \textit{direct} effect. As the video progresses, subsequent frames gradually exhibit a \textit{diminishing connection} to the effect of the initial short action. Note that, the distance of an effect frame being a \textit{measurable} and \textit{observable} aspect, is used as a \textit{proxy} to reflect the \textit{strength} of the \textit{connection} that effect-frame to the action.

\noindent\textbf{Framework.} During training, the network learns to determine the ``distance" (or equivalently, degree of relatedness, DR) of effect-frames from actions. Specifically, each training sample consists of an initial state, a short action clip, and a set of $K$ temporally ordered frames following the action, which form effect-frames. Each of the $K$ effect-frames are assigned a \textit{proxy groundtruth} label which indicates the distance of an effect-frame from the end of the action clip. Specifically, in a series of $K$ temporally-ordered frames, $k$-th frame is assigned a value of $1-k/(K-1)$, where $k$ ranges from 0 to $K-1$. This way, the closest or the most directly-related frame is assigned a proxy-label of 1, while the farthest effect-frame is assigned a proxy-label of 0. The task for the network is to predict these proxy-labels as accurately as possible. Framework is shown in \autoref{fig:ssl_action_spec}. Network parameters are optimized through a regression loss (L2) function. Note that ground-truth labels ($1:1/K:0$) are \textit{artificially} generated and are arbitrary---any other scale can be used as long as it can express how far is an effect-frame from the action clip. We use \textit{unlabeled} videos and \textit{artificially} generated ground-truth to learn representations using this \textbf{self-supervised} pretraining task.  

In the process of solving this task, we hypothesize that the model learns to focus on \textit{intricate} details, such as \textit{identifying} the specific \textit{maneuvers} executed by the athlete and understanding the consequent sequence of \textit{poses}. It also learns to discern how the athlete's position evolves as a direct outcome of these maneuvers. These details have an overlap with the elements of interest in the task of \textbf{action quality assessment} (AQA). Thus, we hypothesize that representations learnt in solving our SSL CATE Effect-Affinity Assessment should help with downstream task of \textit{AQA}. To validate this, the self-supervised model is tested on \textit{action quality score prediction task}, where the model scores \textit{how well} an action was performed. A brief introduction to AQA task and implementation details are provided in the \supplementary.
\section{Experiments}
\label{sec:experiments}

\subsection{Cross-Sample Action Selection}
\label{subsec:full_perf_cmp}
We start by evaluating various baseline approaches on the Action Selection task/dataset.

\noindent\textbf{Baselines.}
As the proposed task is new with no established methods/benchmarks, we design \& explore various baseline methods to connect actions \& effects. Below is a summary of the models; further details in the \supplementary. \textcolor{RoyalBlue}{Dataset \& Codes for all models will be publicly released.}
\begin{itemize}[wide, labelwidth=!, labelindent=5pt, noitemsep,topsep=0pt]
    \item \textbf{Naive baseline:} cosine similarity between the average of the initial and final state features and each of the action options is computed. Action option with the highest similarity score is selected as the correct answer. This model involves no training, pretrained action recognition model is used as the state and action feature extractor.
    \item \textbf{VideoMAE}\cite{videomae_1, videomae_2}\textbf{:}~a model that reconstructs missing parts of video frames to learn video representations.
    \item \textbf{CLIP}\cite{clip}\textbf{:} model learns to understand images and text together by training on a wide variety of internet data to create strong, versatile visual and textual representations.
    \item \textbf{VideoChat2}\cite{videochat2}\textbf{:} strong multimodal foundational model (CVPR 2024) that is simultaneously trained on 20 challenging video understanding tasks. We measure its zeroshot performance.
    \item \textbf{Actions as Transformations}: inspired by \cite{wang2016actions}, we develop a model where a dense network computes a transformation vector from an action clip. The Hadamard product of the initial state vector \& the transformation vector yields the final state vector. Inference: action option producing the highest similarity with the final state feature is chosen as the correct answer. Inference remains same for all baselines.
    \item \textbf{MoRISA:} we leverage a bilinear model \cite{lin2015bilinear} to capture rich interactions between the initial state and the action to generate the final state in feature space.
    \item \textbf{LinSAES:} we learn a linear, disentangled state-action embedding space, where adding action vector to the initial state vector, transitions it to final state vector.
    \item \textbf{Connecting via Swapping:} We develop an encoder-decoder model. Encoders distill State-transformation `codes' (STC) in two ways: 1) from pairs of initial and final states, and 2) from action clips. The decoder learns to construct the final state from the initial state and the STC. By training the decoder to recreate the final state using STCs from both state pairs and action clips, the model learns to understand how actions cause changes and can predict the outcomes of actions.
\end{itemize}

\begin{table}
\centering
\small
\setlength\tabcolsep{10pt}
\begin{tabular}{@{}lc@{}}
\toprule
\textbf{Model}  &  \textbf{Accuracy(\%)} \\ \midrule
Random    &   24.39 \\
Naive   &  47.39    \\
VideoMAE  & 47.21 \\
CLIP & 31.63 \\
VideoChat2 & 38.00 \\
Actions as Transformations    &  47.33     \\
MoRISA   &  52.82      \\
LinSAES    &   54.65         \\
Connecting via Swapping    &     52.25    \\
Analogical Reasoning    &    \textbf{55.20}       \\
Humans & \underline{81.33}  \\ \bottomrule
\end{tabular}
\caption{\textbf{Performance of models on \textit{Action Selection task}.}}
\label{tab:full_perf_cmp}
\end{table} 
\noindent\textbf{Implementation details.} All the models are implemented using PyTorch \cite{pytorch}, and optimized using ADAM optimizer \cite{kingma2014adam} with a learning rate of 1e-4. We train all the models using CATE train set and select the best-performing iteration based on the validation set. The performance of the best model is then verified on the test set. Noting the state-of-the-art performance of Video Swin model \cite{videoswin} on various video understanding tasks, we use a Kinetics pretrained version of it as the backbone feature extractor. Note that CATE task and methods are not limited to video swin model, and future work may use other models. 

\noindent\textbf{Answer evaluation.} 
For the model's answer to be considered correct it has to, from the given options select only one option that matches the ground truth based on the highest similarity. In the scenario where all options return equal similarities, it is deemed to have failed to answer correctly. We use accuracy to evaluate the performance metric.

\noindent\textbf{Quantitative results.} The results are summarized in \autoref{tab:full_perf_cmp}. The Naive approach performs much better than random weights/chance accuracy. VideoMAE performs similar to Naive approach, suggesting that Action Selection and VideoMAE are \textit{different}. Interestingly, foundational models CLIP and VideoChat2 performed inferior to naive model. This maybe because while MLLM/VLM have achieved strong results, they maybe \textit{less} sensitive to finegrained details like \textit{object placement}, \textit{pose}, \textit{motion}, which are crucial in our problem. This also suggests that CATE is a fundamentally \textit{different} task than existing video understanding tasks, and foundational models can \textit{benefit} from including it in their training suite. The Actions as Transformations (AT) approach performs on par with the naive model. However, transformation-based approaches could be made more effective by simplifying the learning process and utilizing \textit{per-element learnable weights} for the \textit{interactions} between the initial state and action vectors (MoRISA). We observe further improvements by learning a linear space for state-action interactions (LinSAES). Although it is a simple approach, it is found to be effective for connecting actions and effects. Connecting via Swapping which involves learning to explicitly tease out or disentangle the `transformation' from actions and states works better than naive and AT approaches; however, it performs slightly inferior to LinSAES. While both approaches can achieve disentanglement in some form, empirically we observe that implicitly disentangling in the process of learning a linear space might be a better alternative than explicit disentanglement via feature partition and swapping. However, the human cognition-inspired Analogical Reasoning model performed the best. This success is likely due to explicitly computing state changes and comparing them with action options.

\begin{figure}[!t]
\small
\centering
\begin{subfigure}[b]{\columnwidth}
     \centering
        \begin{tabular}{@{}ccc@{}}
        {\hspace{0.7cm}Initial State} & {\hspace{0.7cm}Action} & {\hspace{0.3cm}Final State} \\
        \multicolumn{3}{c}{\includegraphics[width=0.9\columnwidth]{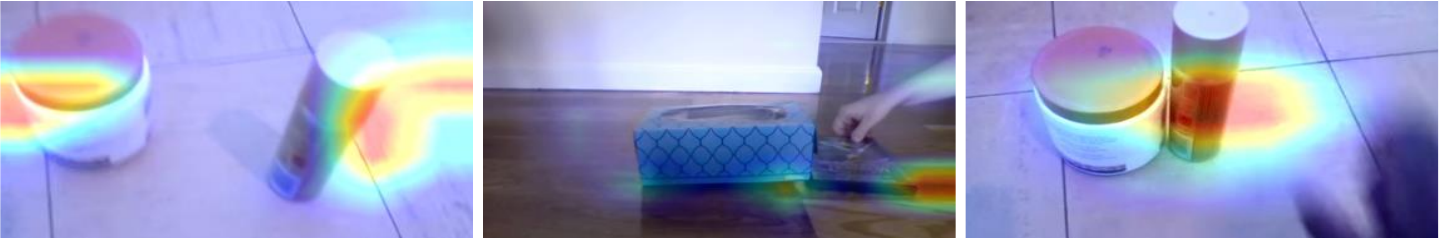}}                               
        \end{tabular}
        \vspace{-0.2cm}
     \caption{}
     \label{fig:attn_viz_a}
 \end{subfigure}
 \\
 %
 \begin{subfigure}[b]{\columnwidth}
     \centering
        \begin{tabular}{@{}c|c@{}}
        {Naive model} & {Analogical Reasoning (AR)} \\ 
        \includegraphics[width=0.45\columnwidth]{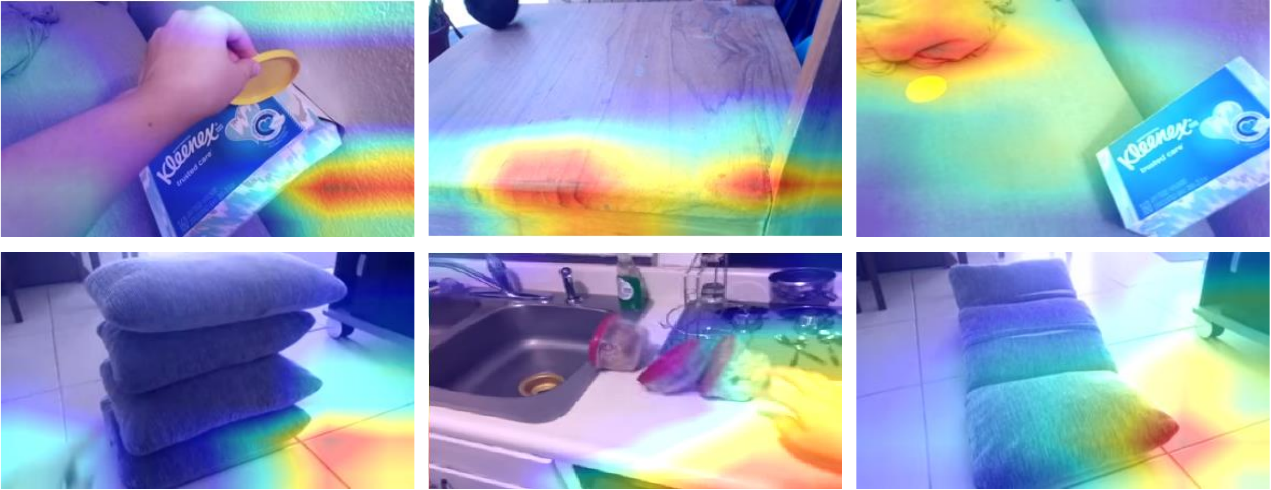}                     & \includegraphics[width=0.45\columnwidth]{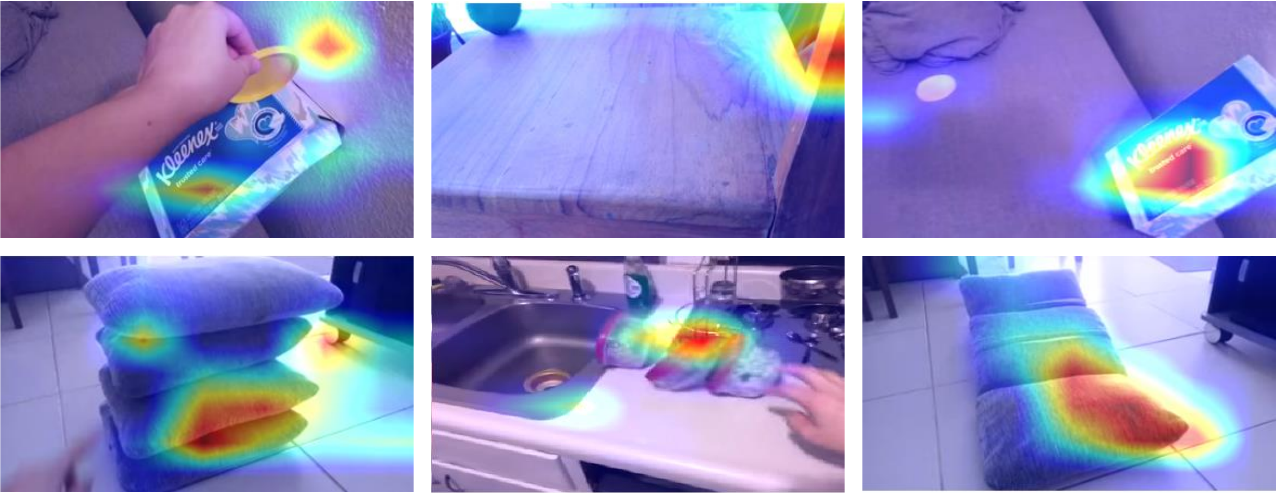}                     \\ 
        \end{tabular}
        \vspace{-0.1cm}
     \caption{}
     \label{fig:attn_viz_b}
 \end{subfigure}
\vspace{-0.7cm}
 \caption{\textbf{Probing where the model pays attention when connecting action \& effects.}. \textbf{(a)} Action is `\textit{putting something close to something}'. Notice how the model tracks the box moved by the person to follow the state change. Note that the states and the action are from different samples. \textbf{(b)} Where the AR model attends vs where Naive model attends. AR model focuses on state changes and driving action and effects; notice the avocado and coin being tracked. Naive model without reasoning module seems to be doing simple matching based on texture. \textit{Zoom in to view better}.}
\label{fig:attn_viz}
\end{figure}

\noindent\textbf{Human baseline.}  
To benchmark our models, we assessed human performance. We had three undergraduate students answer a subset of 100 questions. Their average performance, detailed in \autoref{tab:full_perf_cmp}, significantly exceeded that of all baseline models on this task of Action Selection.

\noindent\textbf{Qualitative results.} 
We analyze where our top-performing Analogical Reasoning model focuses when connecting actions \& effects by extending the GradCAM method \cite{gradcam} (details on our visualization technique are in \supplementary). The visualizations of the model's attention during the Action Selection task are shown in \autoref{fig:attn_viz}.

\subsection{Learning visual representations from unlabeled videos using CATE}

\begin{table}
\small
\centering
\setlength\tabcolsep{5pt}
\begin{tabular}{@{}lccccc@{}}
\toprule
\textbf{Model} & \textbf{Top-1} & \textbf{Top-5} & \textbf{Top-10} & \textbf{Top-20} & \textbf{Top-50} \\
\midrule
VCOP \cite{vcop}  &12.5 &29.0 &39.0 &50.6 &66.9\\
VCP \cite{vcp} &17.3 &31.5 &42.0 &52.6 &67.7\\
MemDPC \cite{memdpc} &20.2 &40.4 &52.4 &64.7 &-\\
VideoPace \cite{videopace} &20.0 &37.4 &46.9 &58.5 &73.1\\
SpeedNet \cite{speednet} &13.0 &28.1 &37.5 &49.5 &65.0 \\
PRP \cite{yao2020video} &22.8 &38.5 &46.7 &55.2 &69.1\\
Temp.Tran. \cite{jenni2020video} &26.1 &48.5 &59.1 &69.6 &82.8\\
TCGL \cite{tcgl} &22.4 &41.3 &51.0 &61.4 &75.0\\
STS \cite{sts} &\underline{39.1} &\textbf{59.2} &\textbf{68.8} &\textbf{77.6} &\textbf{86.4}\\
Ours SSL-CATE       & \textbf{41.5}         & \underline{56.2}         & \underline{64.7}          & \underline{73.5}          & \underline{83.7}     \\
\bottomrule
\end{tabular}
\caption{\textbf{Nearest Neighbor retrieval quantitative results on UCF101.} Best results: \textbf{bolded}; second best: \underline{underlined}.}
\label{tab:res_nn_ret}
\end{table}

\begin{figure}
  \centering
  \small
  \includegraphics[width=\columnwidth]{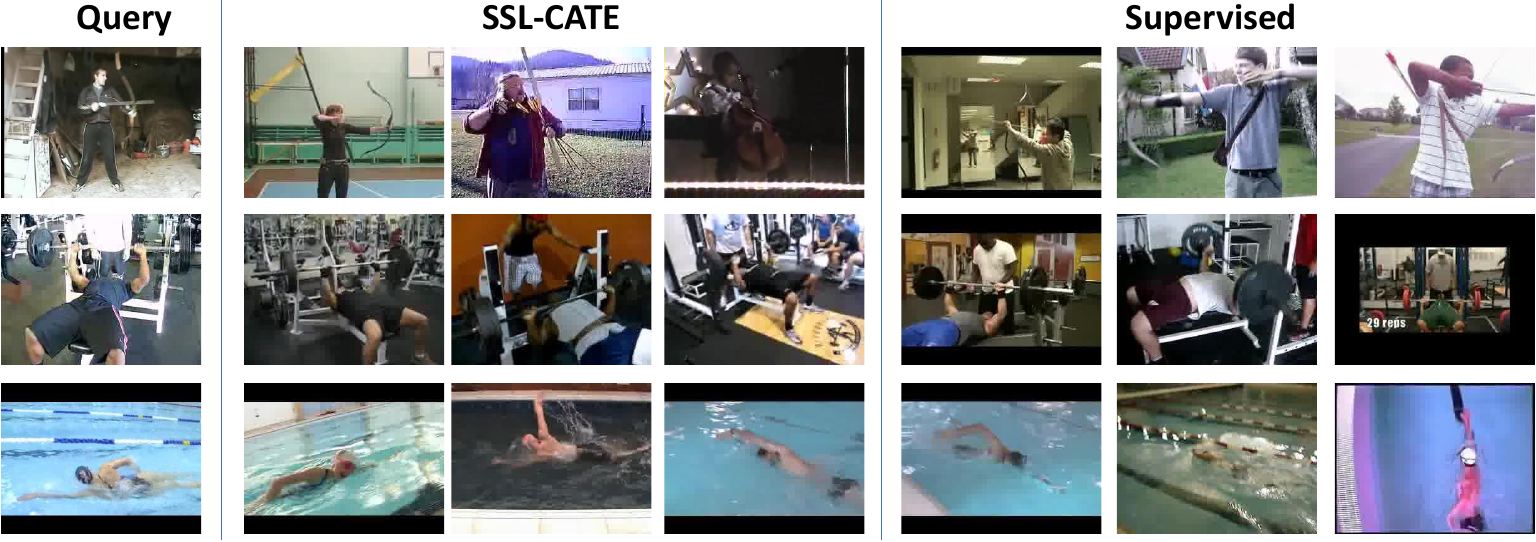}
  \caption{\textbf{Nearest Neighbor retrieval qualitative results.} Visualized Top-3 retrievals on UCF101.}
  \label{fig:nn_ret_ucf}
\end{figure}

After pretraining with the self-supervised action selection task (\autoref{sec:task_formulation_2}) on the UCF101 \cite{ucf101} train set without using any labels, we freeze the backbone (C3D architecture \cite{c3d}). No fine-tuning is performed. The frozen backbone is then used as a feature extractor. Following standard practice in the literature \cite{jenni2020video, videopace}, we conduct a nearest neighbor retrieval experiment. Quantitative results for the nearest neighbor retrieval experiment are presented in \autoref{tab:res_nn_ret}. Our CATE model is compared with other state-of-the-art SSL video representation learning approaches. We observe that SSL-CATE performs the best overall in terms of  Top-1 accuracy, ranking second best in the remaining metrics, only outperformed by STS. These results highlight the significance of CATE in effectively supporting the learning of general action representations.

We further probe into what the model might be learning by retrieving videos similar to the randomly chosen query videos. Results are presented in \autoref{fig:nn_ret_ucf}. We observe that the retrieved results are mostly from the \textit{correct} action class, suggesting that CATE encourages the model to learn the \textit{tracking} of important \textit{human body parts} and \textit{objects}. Upon further investigation of some cases where the retrieval is from the \textit{incorrect} action class, we found that motion patterns in the retrieved videos can potentially result in the \textit{same state change} as the action in the query videos. For example, in \textit{archery} and \textit{playing cello}, \textit{pulling the arrow in archery} and \textit{moving the bow/stick on the cello} involve \textit{similar hand movements} from humans. We further compare the results with a fully-supervised model. Qualitatively, we found the CATE-pretrained results to be \textit{on par} with those from a fully supervised model.

\subsection{SSL Effect-Affinity Assessment for AQA}  

In this case, we train using self-supervision grounded in connecting actions and their effects, as presented in \autoref{sec:task_formulation_3}. Following the literature \cite{roditakis2021towards, fitnessaqa}, we carry out SSL training on the training set of the MTL-AQA dataset \cite{mtlaqa}. After SSL training, we do not perform any fine-tuning; we simply use the backbone as a feature extractor. Features are extracted from 16 uniformly sampled frames, and then a linear regressor is applied on top of them. We compare SSL CATE-Effect-Affinity Assessment with other self-supervised models, and the results are shown in \autoref{tab:app_aqa}. We observe that SSL-CATE Effect-Affinity Assessment self-supervision outperforms other state-of-the-art methods, indicating that CATE-Effect-Affinity Assessment does help in learning features suitable for fine-grained action analysis tasks such as action quality assessment.

\begin{table}
\small
\centering
\begin{tabular}{@{}lc@{}}
    \toprule
        \textbf{Model} & \textbf{Performance}\\
        \midrule
        SSL Action Alignment \cite{roditakis2021towards} & 77.00\\
        SSL Motion Disentangler \cite{fitnessaqa} & 77.63\\
        Ours SSL CATE Effect-Affinity Assessment & \textbf{79.36}\\
        \bottomrule
    \end{tabular}
\caption{\textbf{Performance evaluation on action quality assessment task on MTL-AQA dataset.} Performance metric: Spearman's rank correlation coefficient (\%).}
\label{tab:app_aqa}
\end{table}

\begin{figure}
\small
  \centering
  \includegraphics[width=0.95\columnwidth]{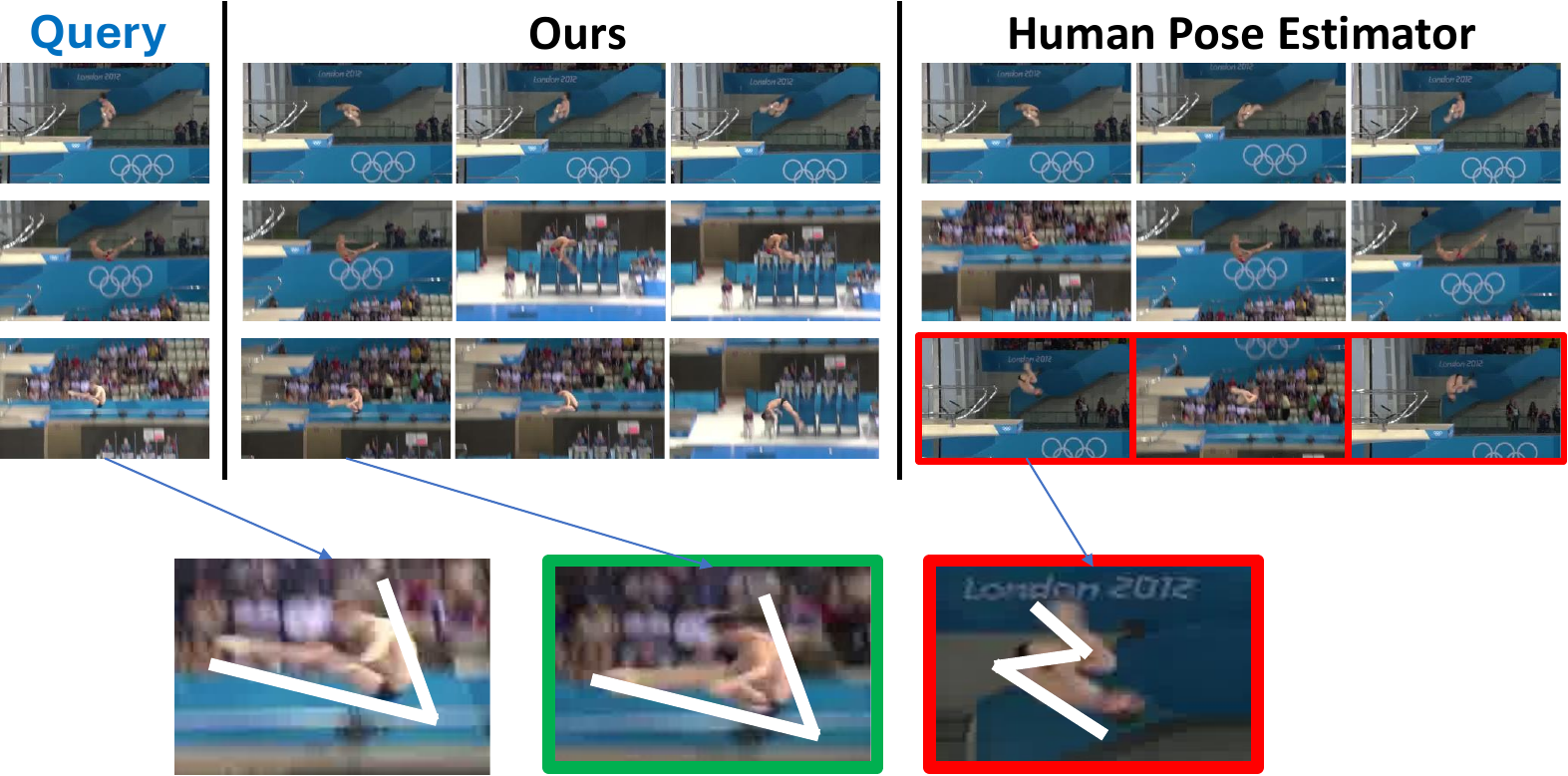}
  \caption{\textbf{Pose retrieval results.} Top-3 retrievals containing diver using our model vs \cite{simplepose}. In this experiment, what we are looking for is if the retrieved results have the athlete in similar pose and orientation as the query frame (left most column). For explanation, we have drawn red bounding box around wrong retrievals. As highlighted with white-colored annotations in zoomed in view, in the \textcolor{red}{wrong retrieval} the athlete is in somersault tuck pose, while the query has the athlete in pike position. In the \textcolor{green}{correct retrieval} as done by our SSL-CATE-Effect-Affinity Assessment, the retrieval results have the athlete in the pike position.}
  \label{fig:res_pose_ret_explained}
\end{figure}

\noindent\textbf{SSL-CATE Effect-Affinity Assessment trained representations learn to encode human pose.}
In this experiment, we further investigate what the SSL-CATE Effect-Affinity Assessment network might be focusing on. Using the trained model from the previous experiment, features are extracted from frames of diving video clips. We then conducted a pose retrieval experiment. Given a frame containing a diver in a certain pose (query pose), the goal is to retrieve frames of divers in similar poses and orientations as the query pose based on feature similarity. We compared the retrieval performance of SSL-CATE trained representations against an off-the-shelf dedicated pose estimation model \cite{simplepose}. Our findings indicate that CATE representations are more effective at retrieving similar poses than the dedicated pose estimation model (\autoref{fig:res_pose_ret_explained}). This better performance suggests that SSL-CATE Effect-Affinity Assessment representations encode fine-grained details like the pose of the person. We hypothesize that this may be because the off-the-shelf pose estimator struggles with \textit{motion blur} and \textit{convoluted poses}, while the CATE-based self-supervision learns \textit{pose-sensitive} and \textit{robust} representations \textit{directly} from \textit{challenging} videos.
\section{Conclusion}
\label{sec:conclusion}
In this work, we explore the concept of connecting actions and their effects (CATE), which has applications in learning from demonstration and task planning. We identify and study two key aspects: Action Selection (AS) and Effect Affinity Assessment (EAA), formalize them, introduce datasets, and develop baseline solutions. While our baselines improve over existing models, a significant performance gap remains between humans and machines on AS. Our experiments show that in solving AS \& EAA, models learn intuitive properties like object tracking and pose encoding without explicit supervision. We demonstrate that CATE can be an effective self-supervised task for learning video representations from unlabeled videos. CATE is a fundamental cognitive process with potential applications in learning from demonstrations, human-machine interactions, and creating adaptive AI tools. Our work lays the foundation for future innovations, inspiring advanced models with broader applications and improved performance.

\paragraph{Acknowledgements.} This research/project is supported by the National Research Foundation, Singapore, under its NRF Fellowship (Award\# NRF-NRFF14-2022-0001).

{\small
\bibliographystyle{ieee_fullname}
\bibliography{egbib}
}

\section*{Appendix}
\label{sec:appendix}
\section{Dataset samples with and without object detections}
For easier viewing, we have provided them in the \textcolor{blue}{accompanying \textbf{Video}}.

\section{Further details on Baseline Methods for Action Selection}
In this section, we provide further details on the various baseline approaches that we design and evaluate. Actions are represented as video clips, while states by very short clips. Unless specified otherwise, we use a backbone model ($\mathcal{E}$) pretrained on a large-scale action recognition dataset to extract features for actions and states. Subsequently, various modules mentioned in the following operate on these features.

\subsection{Naive model}
\label{sec:baseline_naive}
\begin{figure}[!h]
\centering
\includegraphics[width=0.9\columnwidth]{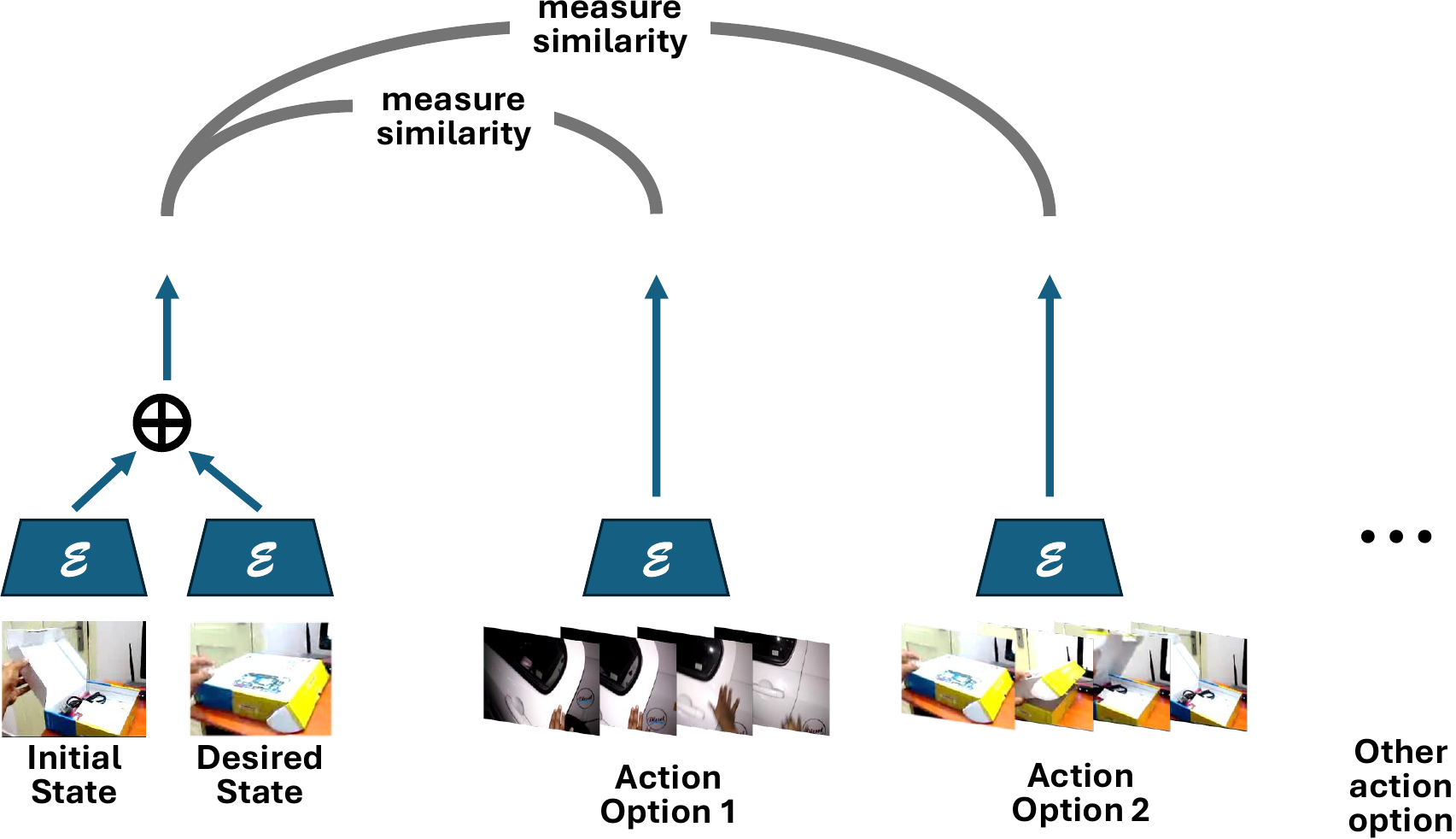}
\caption{\textbf{Naive model.} This model involves no training, but only inference. \textbf{$\bigoplus$} represents averaging operation.}
\label{fig:model_naive}
\end{figure}

Approach is illustrated in \autoref{fig:model_naive}. As a simple baseline, in naive matching the average of the initial and desired or the final state features are matched with the action features using cosine similarity. During inference, the action option with the highest similarity score is selected as the correct answer. Note that, this model involves no training; pretrained action recognition model is used as the state and action feature extractor.

\subsection{Treating Actions as Transformations}
\begin{figure}
\centering
\includegraphics[width=\columnwidth]{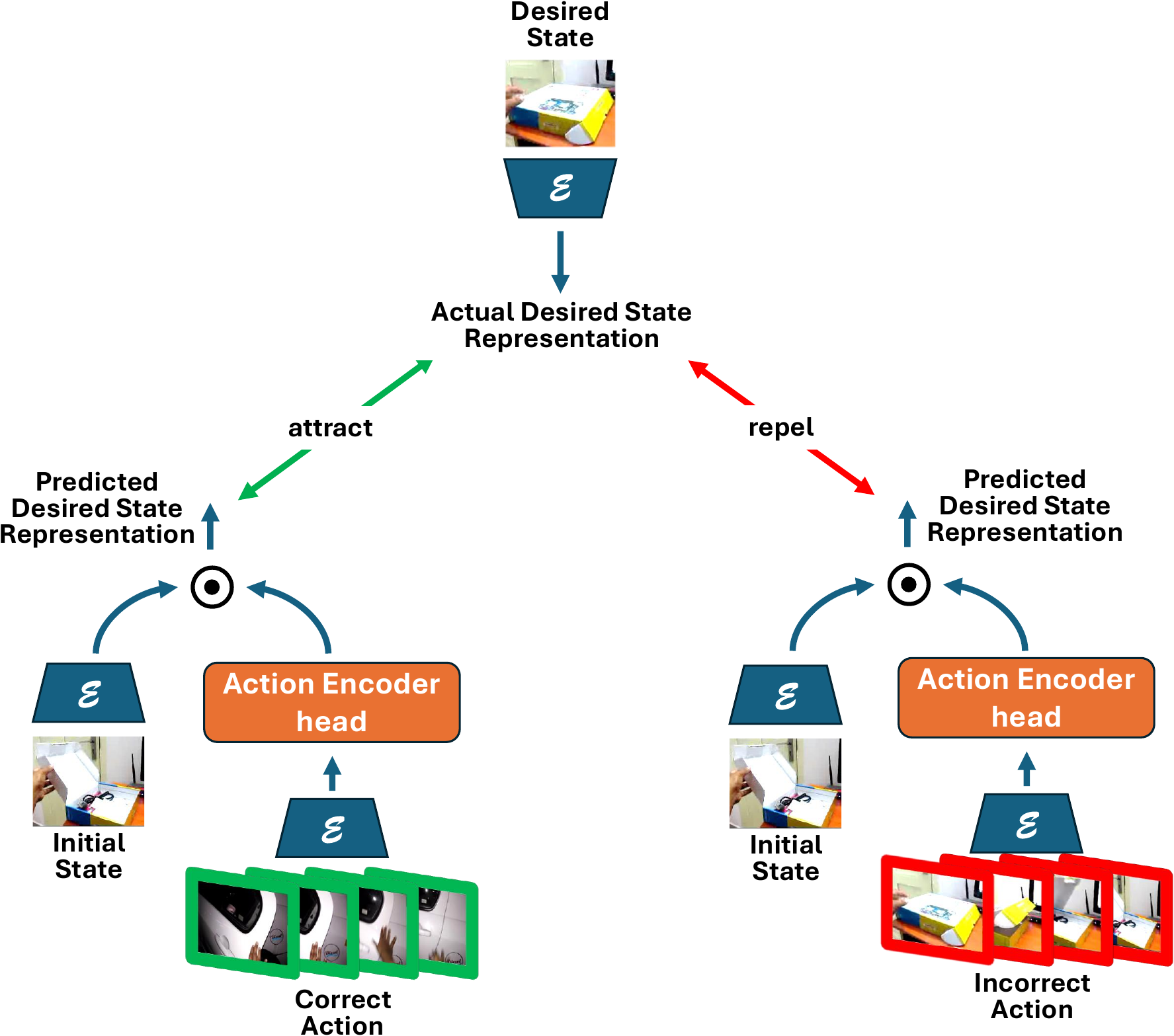}
\caption{\textbf{Actions as Transformations model.} $\bigodot$ represents Hadamard product. Here, we have shown only one incorrect option; but in practice, during training, we consider all the incorrect options.}
\label{fig:model_actions_as_transformations}
\end{figure} 
The approach (in training mode) is shown in \autoref{fig:model_actions_as_transformations}. Actions can be viewed as transformations~\cite{wang2016actions} which when applied to the initial state yields the desired or the final state. While the original approach uses fixed transformations~\cite{wang2016actions}, our adaptation facilitates conditional transformations computed on the fly. We develop a model in which a transformation vector is computed from an action clip using a dense network. This transformation vector ($A$) when multiplied with the initial state vector ($I$) yields the final state vector ($F$). During training, we use contrastive learning~\cite{chen2020simple} to train the parameters of action encoder: Cosine similarity between predicted and actual desired state representations is maximized for the correct action and minimized for the incorrect action. During inference, the action option predicting the final state with the highest similarity to the given final state feature is selected as the correct answer. We term these models as Actions as transformations (AT). 

\begin{figure}
\centering
\includegraphics[width=\columnwidth]{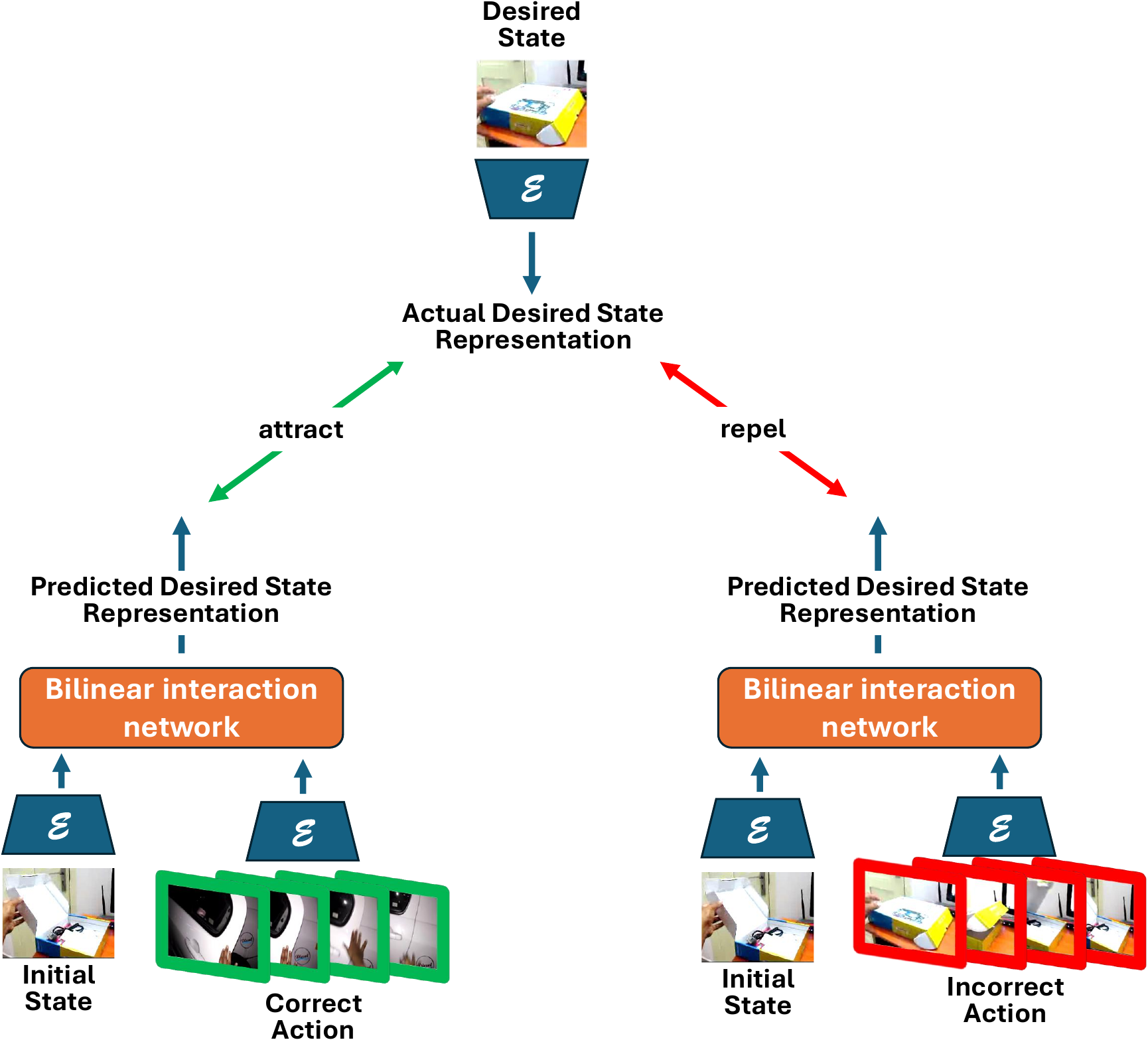}
\caption{\textbf{MoRISA model.}}
\label{fig:model_morisa}
\end{figure} 
\subsection{Modeling rich interactions between initial state \& action (MoRISA)}
Approach (in training mode) is shown in \autoref{fig:model_morisa}. Here a bilinear model \cite{lin2015bilinear} is leveraged to capture rich interactions between the initial state and the action to generate the final state in feature space. Essentially, initial state ($I$) and action representations ($A$) are fed into a bilinear model, which outputs a final state representation ($F$). The parameters of the model are trained using contrastive learning as before and the inference is done similarly to AT model above. 

\begin{figure}
\centering
\begin{subfigure}[b]{\columnwidth}
     \centering
\includegraphics[width=0.7\columnwidth]{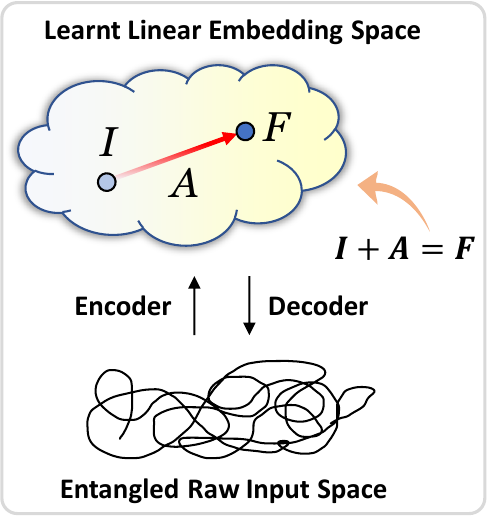}
\caption{\textbf{LinSAES Concept.}}
     \label{fig:model_linsaes_a}
 \end{subfigure}
 \begin{subfigure}[b]{\columnwidth}
     \centering
\includegraphics[width=\columnwidth]{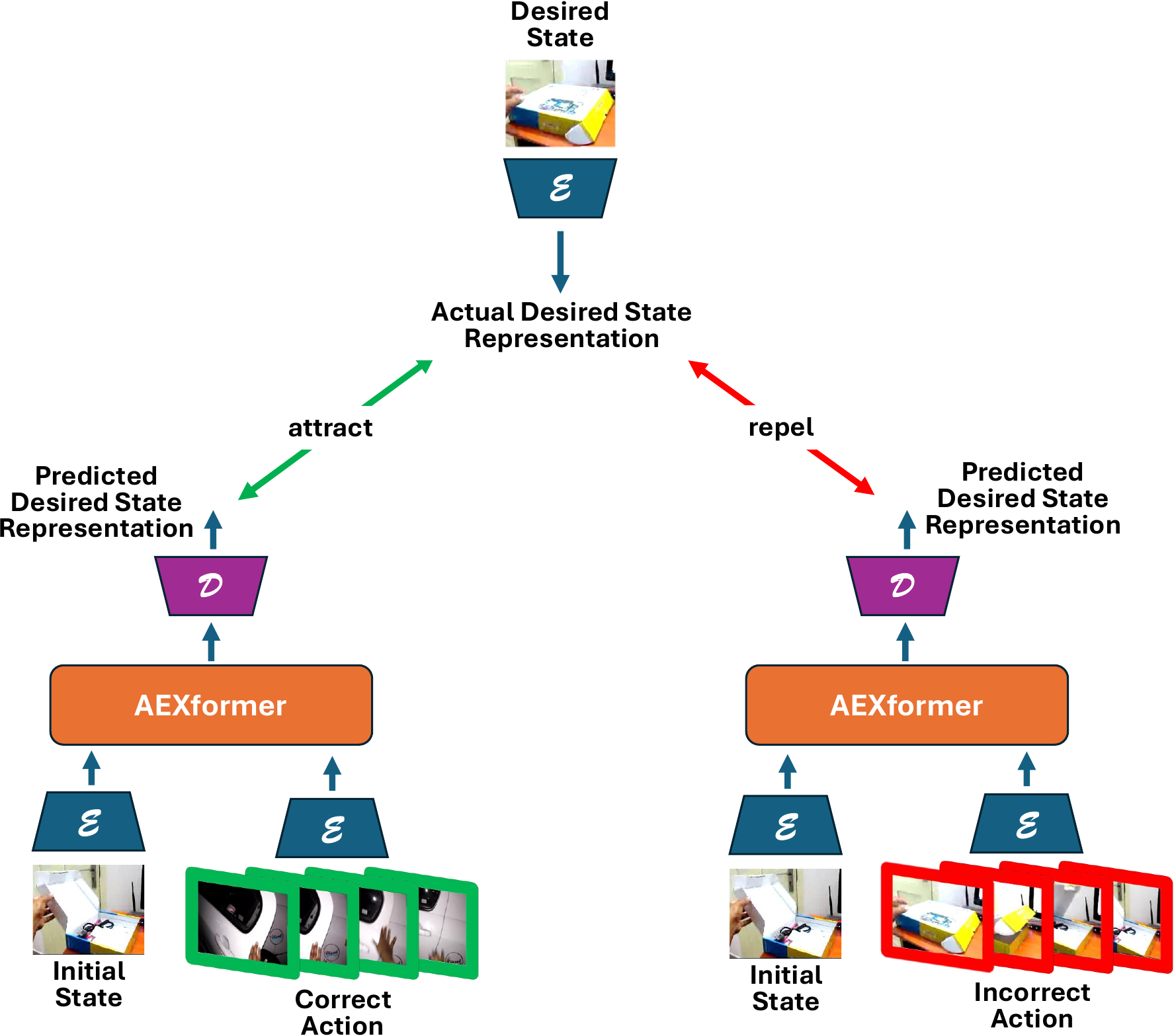}
\caption{\textbf{LinSAES model.}}
     \label{fig:model_linsaes_b}
 \end{subfigure}
\caption{\textbf{LinSAES concept (a) and model (b).}}
\label{fig:model_linsaes}
\end{figure} 
\subsection{LinSAES: Learning Linear State-Action Embedding Space}
Concept of this approach is shown in \autoref{fig:model_linsaes_a}; and the framework (in training mode) is shown in \autoref{fig:model_linsaes_b}. In this framework, actions and initial states undergo transformation into a linear embedding state-action representation space through a Transformer encoder (we term it as AEXformer in \autoref{fig:model_linsaes_b}). The hypothesis suggests that within this \textit{linear}, \textit{disentangled} space, \textit{adding} an action vector ($A$) to the initial state representation ($I$) enables a transition to the final state in the representation space ($F$) (illustrated in conceptual diagram in \autoref{fig:model_linsaes_a}). The Transformer encoder processes a sequence of initial state and action representation vectors, and the resulting output is decoded by a linear decoder ($\mathcal{D}$) to produce the final state feature vector. During training, all model parameters are optimized end-to-end with the objective of aligning the resultant final state representation with the ground truth final state representation. Optimization and inference process remains the same as AT model. We call this model LinSAES. 

\begin{figure}
\centering
\includegraphics[width=\columnwidth]{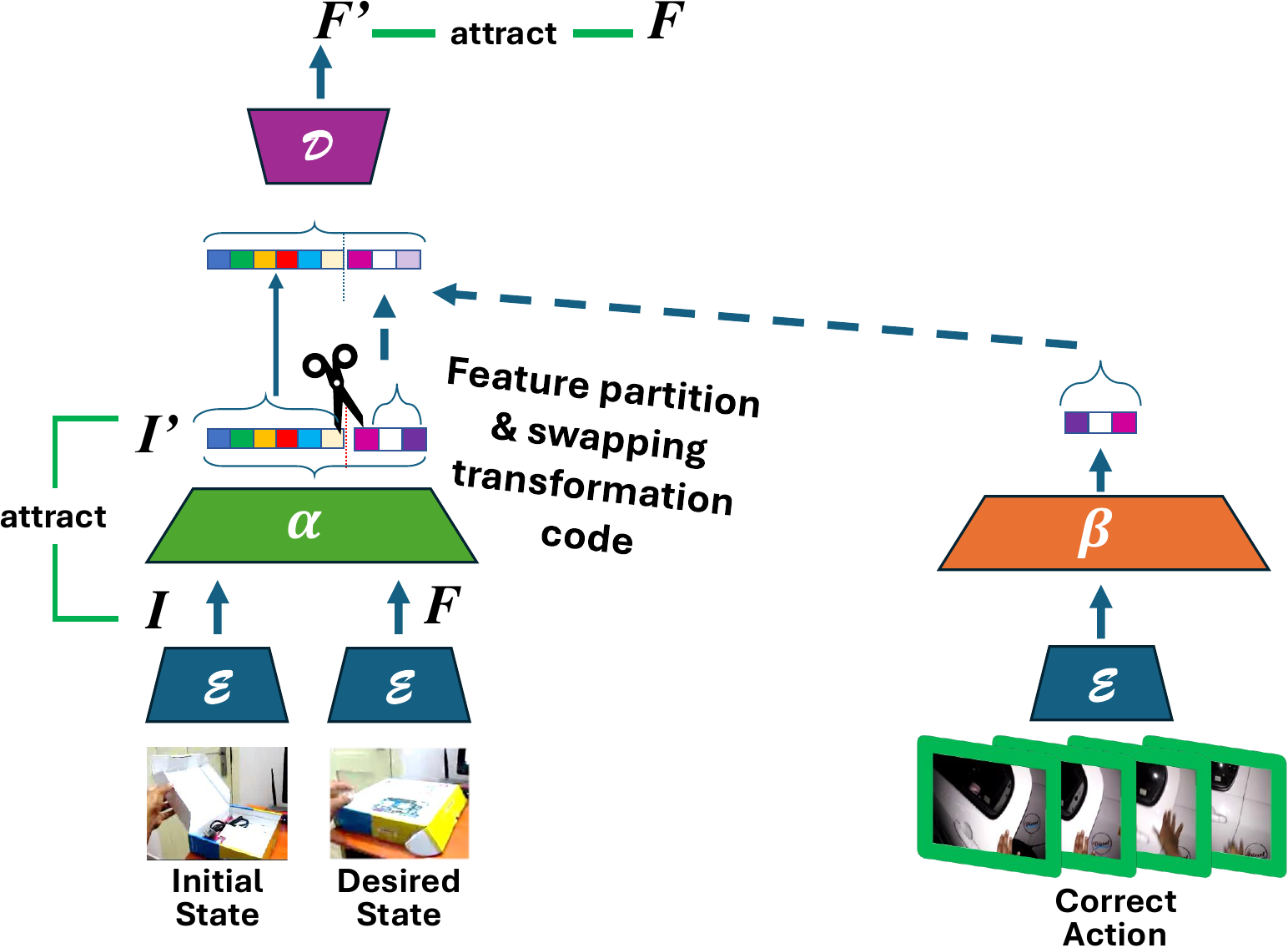}
\caption{\textbf{Connecting via Swapping.} Here, we have not shown using of incorrect action for better explanation and avoiding confusion. However, in practice, we leverage incorrect actions.}
\label{fig:model_connecting_via_swapping}
\end{figure} 
\subsection{Connecting via Swapping}
We will start by presenting method overview, followed by details. Approach (in training mode) is illustrated in \autoref{fig:model_connecting_via_swapping}. 
\paragraph{Method overview:} all model parameters (encoders ($\alpha, \beta$) and decoders ($\mathcal{D}$)) are trained end-to-end by enforcing the following. 1) \textbf{Transformation Encoding.} The initial and final states are transformed into an initial state and a corresponding state-transformation code through an encoder. This transformation code captures the changes or effects between the initial and final states. 2) \textbf{Reconstruction using Decoder.} Given the initial state and the state-transformation code, a decoder reconstructs the final state. This decoder essentially learns to generate the final state based on the initial state and the transformation code. 3) \textbf{Obtaining Transformation Code from Cross-sample Actions.} Additionally, a state-transformation code can be obtained from the action (but from another sample) occurring between the initial and final states. This is done through another encoder that distills the transformation code from the action. Note that, during training the final state is reconstructed using both transformation codes---obtained from states and action. By obtaining equivalent state-transformation codes both from the states directly and from the actions, the model learns to connect actions to their effects. The decoder then utilizes this transformation code to reconstruct the final state from the initial state, effectively capturing the impact of actions on the state transition in the video data.
\paragraph{Method details:} To explicitly isolate a $n$-dimensional ``transformation ``code", we pass the initial and final states (each represented by $m$-dimensional features) through an encoder $\alpha$, which outputs a $(m+n)$-dimensional vector. We enforce the condition that the first $m$ elements of this vector represent the initial state and the remaining $n$ elements represent the transformation ``code". Now, using another encoder ($\beta$), we distill a $n$-dimensional transformation code from an action option. We swap the transformation code obtained using states with that obtained from the action option. A decoder ($\mathcal{D}$) then reconstructs the final state from the extracted initial state and swaps the transformation code. All encoders and decoders are trained by enforcing that the extracted initial state be similar to the ground truth initial state and the reconstructed final state be similar to the ground truth final state in case the transformation code was coming from the correct action option (minimize the similarity if the transformation code was from incorrect action option). This approach is inspired by \cite{fitnessaqa}. However, while their approach disentangles human pose and appearance, our approach extends it to disentangle state-transformation for linking actions and their effects. 

\subsection{CLIP}
For this baseline, we adopt the previously discussed naive approach \autoref{sec:baseline_naive}. In this case, we use the CLIP \cite{clip} representations for states and actions. Further, to obtain a clip-level representation for action-clips, we average the frame-level representations. Inference process remains as the Naive approach.

\subsection{VideoChat2}
VideoChat2 \cite{videochat2} is powerful video understanding foundational model. It is trained on 20 challenging video understanding simultaneously. We evaluated various ways to adopt VideoChat2 for our problem with the objective of maximizing the performance on Action Selection. We obtained best results with the following approach. First, we provide the initial and the final states to the model alongwith the four action classes in language format, ask to identify the action taking place. Second, we ask the model to identify the action class (from the four choices in language format---these are nothing but the class names of the four action options) taking place in the correct action-clip. Then, we match the classes obtained from states and action-clips. If there is a match and the predicted class is correct, we consider that the model got the answer, else, we consider the answer of the model as wrong.

\subsection{VideoMAE}
Mask autoencoding (MAE) might seem similar to Action Selection, but there is a fundamental difference between them. MAE randomly masks patches without any explicit attention to initial and final states, whereas Action Selection particularly involves connecting actions and desired state changes. Nonetheless, VideoMAE has achieved strong results in video understanding. Therefore, for completeness, we consider VideoMAE as one of the baselines. By design, VideoMAE cannot be directly used for our task, so we adopt the previously discussed Naive approach, but using VideoMAE as the backbone for Action Selection task.

\section{Introduction to Action Quality Assessment (AQA) and how it is connected to Effect-Affinity Assessment}

\noindent \textbf{Definition of AQA.} AQA is a fine-grained action analysis task, where the models try to assess how well an action was performed. Judging during Olympics diving can be a classic example of AQA. AQA requires paying attention to very fine-grained details of action. For example, it involves analyzing how tight was the athlete's form during somersaults, how close the athlete's feet were during twisting motion in air, was the athlete under-/over-rotated during entry into the water, how high the athlete jumped during take-off, \etc. Taking these into consideration, a score is given out to indicate how good the athletes' performance was or whether the rules of the competition were followed.

\noindent \textbf{Effect-Affinity Assessment and AQA.} In our formulation, Effect-Affinity Assessment involves discerning between very nearby effects, and determining how far is an effect-frame from the action applied on an initial state. To solve this task, the model learns to pay attention to fine-grained details like what maneuvers the athlete applied/executed, and how that would result in what series of poses, or how the athlete's position would evolve as a result of their maneuvers. These details have an overlap with the elements of interest in AQA. Therefore, we hypothesize that representations learnt in solving our SSL CATE Effect-Affinity Assessment should transfer well to AQA task.

\section{Action-Effect Joint Attention Visualization}
We use the best performing Analogical-Reasoning model for \textit{attention visualization}, employing a modified GradCAM \cite{gradcam} to generate joint attention heatmaps over states \& actions. Specifically, we backpropagate the dot-product of state-change vector \& the action vector through the initial, final, action branches into visual input space.

\section{Further Action Selection performance analysis}
\begin{figure*}
    \centering
    \includegraphics[width=\textwidth]{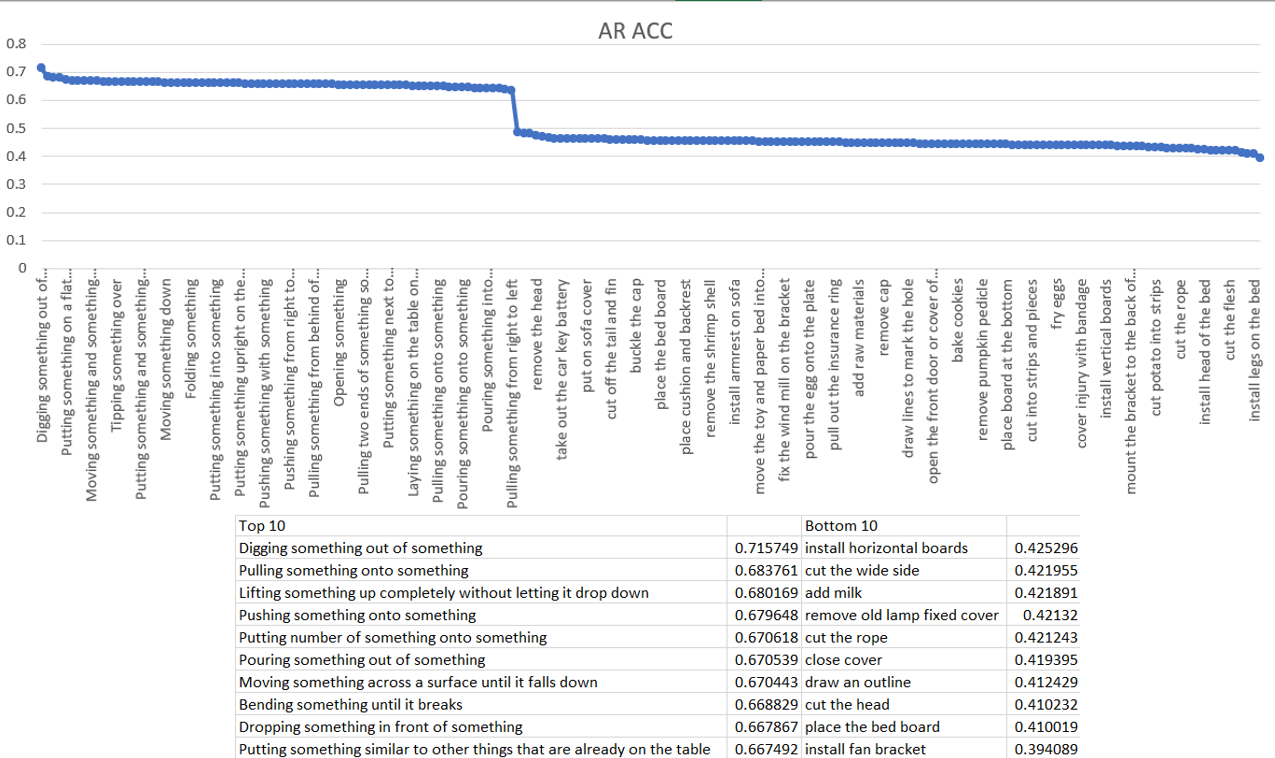}
    \includegraphics[width=\textwidth]{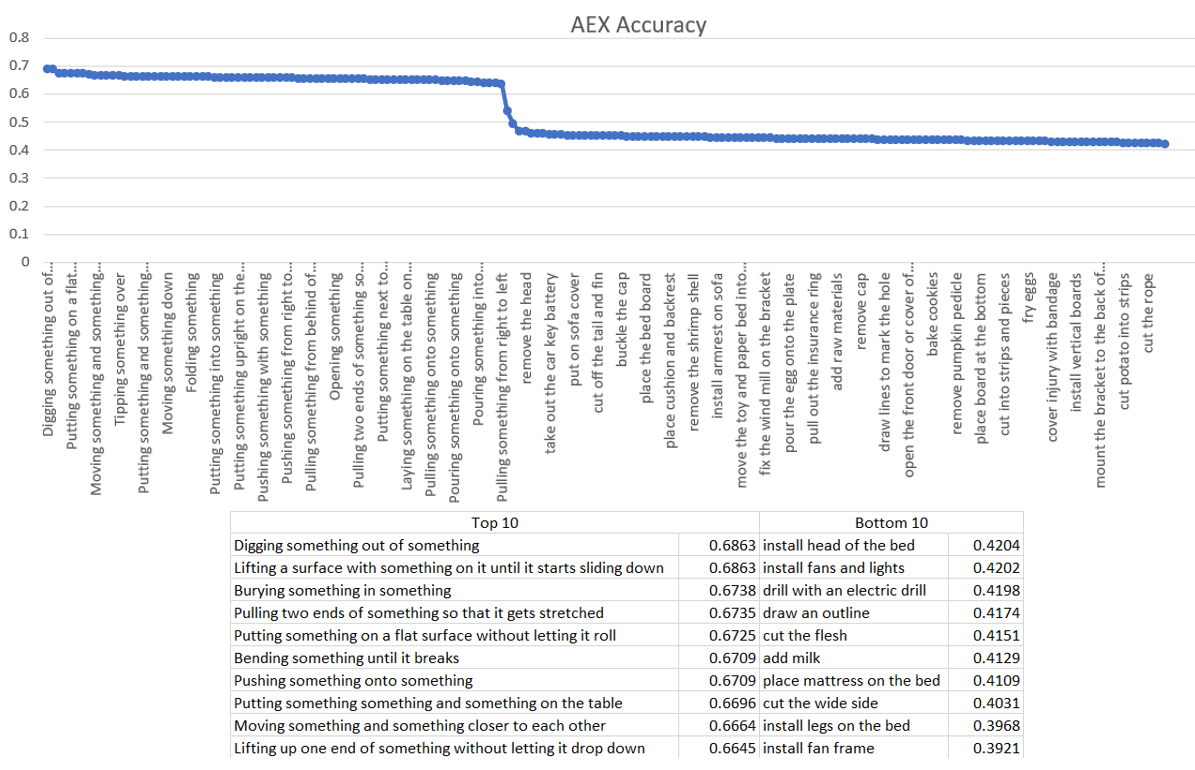}
    \caption{\textbf{Classwise accuracies from Analogical Reasoning (AR) LinSAES (AEX) models.} Easiest and most difficult classes are listed below.}
    \label{fig:classwise_accu}
\end{figure*}

We have also provided the classwise accuracies from two of the best performing models: 1) Analogical Reasoning model; 2) LinSAES (learning linear state-action embedding space) in the \autoref{fig:classwise_accu}. In both cases, we observed a similar trend that the action classes on which the models were most accurately connected actions and effects were simple and single-step movements. Action classes where the models struggled the most were composite and complex involving multiple steps or sub-actions. These composite actions may require the model to understand not only individual actions but also their sequential dependencies and temporal relationships. The increased complexity can make it more challenging for the model to accurately predict the correct sequence of actions to achieve the desired outcome. The easiest classes to connect were derived from SSv2 dataset \cite{ssv2} representing common tasks or actions one might encounter in daily life or simple mechanical tasks. These actions focus on the manipulation and movement of objects in various ways, often on flat surfaces. Most difficult to connect actions were from COIN dataset \cite{coin}, often associated with particular activities such as cooking, construction, or assembly.  These actions involve more specific and varied tasks such as cutting, filling, inserting, and cooking, indicating a wider scope of activities.

\end{document}